\documentclass{article}
\usepackage{ijcai24}

\usepackage{times}
\usepackage{soul}
\usepackage{url}
\usepackage[hidelinks]{hyperref}
\usepackage[utf8]{inputenc}
\usepackage[small]{caption}
\usepackage{graphicx}
\usepackage{amssymb}
\usepackage{amsmath}
\usepackage{amsthm}
\usepackage{booktabs}
\usepackage{algorithm}
\usepackage{algorithmic}
\usepackage[switch]{lineno}
\usepackage{paralist}

\usepackage{amsmath,amsthm,amssymb,multirow,paralist,mathrsfs,amsfonts,dsfont,tikz}

\usepackage{multirow} 
\usepackage{xcolor}

\newtheorem{theorem}{Theorem}
\newtheorem{lemma}{Lemma}

\newtheorem{remark}{Remark}
\let\oldremark\remark
\renewcommand{\remark}{\oldremark\normalfont}

\usepackage{verbatim}

\usepackage{adjustbox}

\def\tr{\text{tr}}

\def\calY{\mathcal Y}

\def\calD{\mathcal D}
\def\calZ{\mathcal Z}
\def\calP{\mathcal P}

\def\E{\mathbb E}
\def\P{\mathbb P}
\def\R{\mathbb R}
\def\I{\mathbb I}

\title{Streamflow Prediction with Uncertainty Quantification for Water Management: \\ A Constrained Reasoning and Learning Approach}

\author{
Mohammed Amine Gharsallaoui 
\and Bhupinderjeet Singh
\and Supriya Savalkar
\and Aryan Deshwal
\and Yan Yan
\and Ananth Kalyanaraman
\and Kirti Rajagopalan
\And Janardhan Rao Doppa \\
\affiliations
Washington State University, Pullman, WA, USA
\emails
{ \{m.gharsallaoui, bhupinderjeet.singh, supriya.savalkar, aryan.deshwal, yan.yan1, ananth, kirtir, jana.doppa\}@wsu.edu}
}

\begin{document}

\maketitle

\begin{abstract}
    Predicting the spatiotemporal variation in streamflow along with uncertainty quantification enables decision-making for sustainable management of scarce water resources. Process-based hydrological models (aka physics-based models) are based on physical laws, but using simplifying assumptions which can lead to poor accuracy. Data-driven approaches  
    offer a powerful alternative, but they require large amount of training data and tend to produce predictions that are inconsistent with physical laws.  
    This paper studies a {\em constrained reasoning and learning (CRL) approach} where physical laws represented as logical constraints are integrated as a layer in the deep neural network. 
    To address  small data setting, we develop a theoretically-grounded training approach to improve the generalization accuracy of deep models. For uncertainty quantification, we combine the synergistic strengths of Gaussian processes (GPs) and deep temporal models (i.e., deep models for time-series forecasting) by passing the learned latent representation as input to a standard distance-based kernel. Experiments on multiple real-world datasets demonstrate the effectiveness of both CRL and GP with deep kernel approaches over strong baseline methods.
\end{abstract}

\vspace{-2.0ex}

\section{Introduction}

\begin{figure*}[t!]
    \centering 
    \includegraphics[scale=0.45] {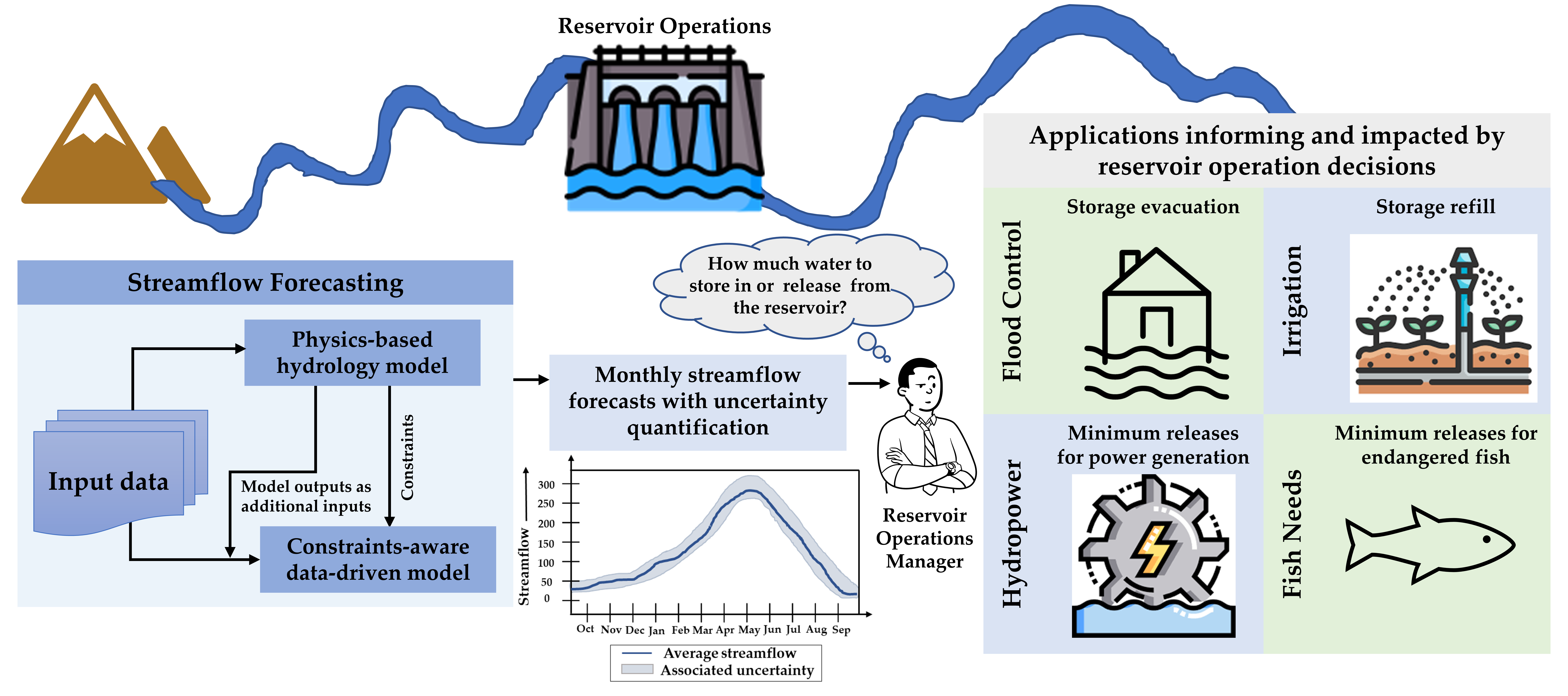}
    \caption{Overview of our constrained reasoning and learning approach for streamflow forecasting to enable water management decisions.}
    \label{fig:problemsetup}
\end{figure*}

Streamflow is fundamental to meeting societal needs including food and energy security, and environmental health. 
In ``fully appropriated'' Western US watersheds---where every drop of water is associated with a beneficial use under a legal water rights system---an accurate representation of the spatio-temporal variation in streamflow is essential for sustainable management of scarce water resources. 
{A typical decision context in the snow-dominant Western US is reservoir operations leading into the spring and summer months when most of the streamflow is generated ({\em high streamflow months} from March through July which is the focus of our study). 

The reservoirs need to be operated to manage competing objectives based on expectations of upcoming streamflow. Typical objectives include meeting flood control,  irrigation demands for the season,  and hydro power generation needs (see Fig. \ref{fig:problemsetup}). One concrete example of the conflicting objectives is as follows. The reservoirs need to be evacuated enough to capture upcoming streamflow and avoid flood risk. However, this cannot be at the expense of not being able to fill the reservoir before the irrigation season starts. Having a good forecast of expected flows in high flow months, and reliable uncertainty bounds around it can help optimize storage and release decisions to meet all objectives. Moreover, different decisions will involve utilizing uncertainty bounds in different ways. When flood risk needs to be minimized, the reservoir manager will likely utilize flows at the upper bound of the uncertainty interval in water release decisions. In contrast, while focusing on meeting irrigators' needs, the reservoir manager will likely make storage decisions based on the lower bound of expected streamflow. Under the auspices of a federal grant, this project team has been interacting with United States Bureau of Reclamation (USBR)---the federal agency tasked with managing reservoir operations for all reservoirs in the Western United States where water for irrigation is a key operational priority along with a multitude of other uses (see Fig. \ref{fig:problemsetup}). Therefore, this work has a real potential for impact at the U.S. national scale. }

Process-based hydrological models  (aka physics-based models) that take weather, soil, and other inputs to translate precipitation to streamflow using biophysical representations of the hydrological process are commonly used to estimate streamflow. These models have been known to yield best results when calibrated to individual watersheds but they continue to struggle in regional-scale modeling across multiple watersheds because hydrological response is heterogeneous and watershed-specific \cite{kratzert2019toward}.

Data-driven (aka deep learning) models can capture this heterogeneity and push the frontiers in learning universal, regional, and local hydrological behavior \cite{kratzert2019toward,peters2017scaling,Water} while increasing streamflow prediction accuracy.  However, this learning methodology has to maintain the basic laws  governing hydrological processes (e.g., conservation of mass for water balance) and capture the essence of important intermediary processes (e.g., evapotranspiration) that translate precipitation to streamflow. Since regional-scale water management decisions are made on ``seasonal'' time scales, we seek to make streamflow prediction for high-flow months to allow a targeted impact. As a result, we lack large amounts of training data required by the deep learning methods---e.g., even with 25 years of monthly data, there are only 125  samples.

Furthermore, in watersheds with competing water uses, when streamflow predictions go wrong, their societal impacts are significant---including losses for agriculture and the rural economy given less-than-ideal drought management; and negative impacts on endangered fish species due to higher/lower flows than expected, e.g., the 2015 drought year resulted in $\sim$\$0.7 billion agricultural revenue loss in the U.S. Washington State \cite{McLain2015}.  
Therefore, quantifying uncertainty under which water managers make these decisions is critical for public trust, conflict resolution, and informed decision making. 

\vspace{0.75ex}

\noindent {\bf Challenges.} There are three key technical challenges: 
{\bf 1)} Predicting outputs which are consistent with the physical laws; {\bf 2)} Training predictive models from small data which generalize well to unseen data; and {\bf 3)} Quantifying the uncertainty in terms of a prediction interval such that the ground-truth output lies in this interval in most cases. 

\vspace{0.75ex}

\noindent {\bf Contributions.} This paper's key contribution is a principled approach for streamflow prediction and uncertainty quantification to address the above-mentioned three challenges.

Prior work has attempted to address the first challenge using physics-guided training or physics-informed neural networks \cite{TGDS,PINNs} where the key idea is to penalize the deep model when the predictions are not consistent with the physical laws. This approach was shown to improve over the pure data-driven approach on predicting depth-specific lake water temperatures \cite{Lake3,Lake2,Lake1} and other use-cases \cite{Assembly}. However, this approach and similar methods \cite{NSL-Survey,mass-lstm,sit2021short,xiang2022fully} do not guarantee that {\em every} test-time prediction is consistent with the domain knowledge.  
To overcome this drawback, as our first contribution, \emph{we study a new constrained reasoning and learning (CRL) approach using a semantic probabilistic layer (SPL)} \cite{SPL}. SPL compiles the domain knowledge specified in the form of logical constraints into a tractable circuit representation \cite{PC1} and adds it as a replacement for predictive layers in deep networks for efficient end-to-end training. We empirically show that CRL approach with SPL instantiation uses the domain knowledge to improve the predictive accuracy over baseline methods.

To address the small data challenge, \emph{we propose a novel {\em importance-weighted (IW)} training approach to improve the generalization accuracy of predictive models learned over the latent representation from the deep temporal model}. 

The key principle is to assign importance weights directly proportional to the errors on data samples during the training process, and to compute the weighted gradients using the data samples for gradient-based training. We provide theoretical analysis to justify the IW training approach. Our experiments show that IW training further improves the predictive accuracy over the standard training. Our IW algorithm and its theoretical analysis is specific to the small data challenge unlike prior work \cite{IW-ICLR2021}.

For uncertainty quantification (UQ), {\em we study a principled approach by combining the synergistic strengths of Gaussian processes (GPs) and deep temporal models}. Deep neural networks are known to extrapolate over-confidently \cite{BayesianDL}. 
On the other hand, GP models are gold standard \cite{williams2006gaussian} for UQ providing principled Bayesian uncertainty estimates, unlike existing ad-hoc techniques \cite{pi3nn}, but have difficulty scaling to high-dimensional structured inputs (e.g., time-series). Deep kernel learning (DKL) \cite{wilson2016deep} is a recent promising  technique  towards handling this drawback while maintaining the good UQ properties of a GP.  
Our experiments demonstrate that GP based approach with deep kernel produces prediction intervals with high coverage (i.e., ground truth output lies within the interval). 
We make our code and datasets available to the AI community at \url{https://github.com/aminegha/StreamPred} to catalyze research in this important problem space.

\vspace{-1.5ex}
\section{Background and Problem Setup}
In streamflow prediction, utilizing graph representations is an efficient method for illustrating watersheds. Within a defined geographical expanse, a watershed represents the entirety of land that contributes water to a shared outlet, which might be a river, lake, or ocean. Within the watershed, a river network showcases the flow paths and connections of water in that region. A watershed is subdivided into grids, representing discrete small sections that facilitate the analysis and modeling of factors such as water movement. In the graph representation, the \emph{nodes} represent the grids within the watershed, and the \emph{edges} represent the flows between them. Each node has a set of \emph{environmental} features (including variables related to precipitation and temperature). 
Given this rich graph representation, our objective is to predict the streamflow at the watershed outlet at fixed time intervals. Since water management decisions are made seasonally, {\em we aim to make streamflow predictions for high-flow months (March to July)}.

\vspace{0.5ex}

\noindent {\bf Domain Knowledge.} The water balance equation (WBE) is a fundamental concept in hydrology: the amount of water in a system should be balanced between the inputs (e.g., precipitation), outputs (e.g., evapotranspiration, streamflow), and changes in storage (e.g., soil moisture or snowpack).
It is given by $P \geq ET + S$, where $P$ is precipitation, $ET$ is evapotranspiration, and $S$ is streamflow. WBE is not an equality because all components of the water balance are not present here and  holds  only at an annual aggregated level.

\vspace{0.75ex}

\noindent {\bf Problem Setup.} For a given watershed, we consider a sequence of graphs over discrete time steps denoted as $G_1, G_2,\cdots, G_T$, where each graph $G_t$ represents a directed graph with a set of nodes $V$ and a set of edges $E \subseteq V \times V$. At each time step, the corresponding features matrix $X_t \in \mathbb{R}^{n \times m}$, where $n=|V|$  and $m$ is the number of features. The graph $G_t$ is also characterized by an adjacency matrix $A \in \mathbb{R}^{n \times n}$, with the property that $A_t$ remains invariant since grid connections are static. At each (monthly) time step $t$, we also have the target variable $y_t$ which is the streamflow at the watershed \emph{outlet} node. Given a training data of time-series $\mathcal{D} = \{(X_t, y_t)\}_{t=1}^{T}$ and domain knowledge $\mathcal{K}$ in the form of water balance equation, we have two goals:
\begin{compactenum}[a)]
\item To learn a predictor $F$ that can make accurate predictions consistent with the knowledge $\mathcal{K}$ for unseen inputs $X_{t^\prime}$ where $t^\prime > T$ ({\em high-flow months} only).
\item  To get uncertainty estimates in the form of a prediction interval $[lb, ub]$ for $X_{t^\prime}$, where the true output $y_{t^\prime} \in [lb, ub]$ with a high probability. 
\end{compactenum}

\section{Technical Approach}

In this section, we outline our overall methodology shown in Algorithm \ref{alg:nsl}  to address the three technical challenges, namely, predictions consistent with domain knowledge, small training data, and uncertainty quantification. 
\subsection{Constrained Reasoning and Learning}
\label{sec:NSL}
\vspace{1.0ex}

\noindent {\bf Recurrent Convolutional Graph Neural Network.} We employ a recurrent convolutional graph convolutional neural network (RCGNN) \cite{GCRNN-2016,sit2021short,xiang2022fully} to solve the streamflow prediction task. RCGNN is a generalization of classical graph convolutional network that incorporates the recurrent nature of RNN to capture temporal dependencies in graph-structured data. 
Notably, in graph embedding learning, for a given node, we integrate information from both preceding time steps and adjacent  river segments (i.e., edges). 

The RCGNN model establishes transition connections for the derived latent representation using a recurrent cell configuration. In the scope of this study, we employ the Gated Recurrent Unit (GRU) cell, known for its prowess in capturing sequential patterns and dependencies in graph-structured data. 
For time step $t$, given a graph represented by its adjacency matrix $A$ and its input feature matrix $X$, a graph convolutional GRU layer is parameterized with multiple weight matrices $W$'s, $U$'s and bias terms $b$'s, and starts by computing the reset gate $r_t$ and updates gate $z_t$ from time step $t-1$:
\vspace{-0.05in}
    \begin{align*}
        & r_t = \sigma(W_r \cdot X + U_r \cdot (A \cdot H^{(t-1)}) + b_r) \\
        & z_t = \sigma(W_z \cdot X + U_z \cdot (A \cdot H^{(t-1)}) + b_z) ,
    \end{align*} 
    where $H^{(t-1)}$ is the hidden state of the previous time step $t$, while $\sigma$ denotes the sigmoid activation function. 
Next, it computes candidate hidden state (\(h_t'\)):
\vspace{-0.05in}
    \begin{align*}
    h_t' = \tanh(W_h \cdot X + U_h \cdot (r_t \odot (A \cdot H^{(t-1)})) + b_h) .
    \end{align*} 
    Here, \(\odot\) represents element-wise multiplication.
Using $z_t$ and $h_t'$, we update the current hidden state $H^{(t)}$ as follows:
\vspace{-0.05in}
    \begin{align*}
    H^{(t)} = z_t \odot H^{(t-1)} + (1 - z_t) \odot h_t' .
    \end{align*}
Therefore, the learnable weight matrices of the model are specifically denoted above by $W_r \text{, } U_r\text{, } W_z\text{, } U_z\text{, } W_h \text{ and } U_h$, and trainable bias terms are $b_r\text{, } b_z \text{ and } b_h$.   
For the subsequent output layers, two distinct activation functions are employed alongside two linear layers. For the activation function, we use LeakyRelu to introduce non-linearity. 
We build our RCGNN model to learn the graph embedding according to the above architecture, denoted by $f(X)$.

\vspace{0.75ex}
\noindent {\bf Semantic Probabilistic Layer.} To address the challenge of maintaining domain knowledge consistency, the constrained reasoning and learning (CRL) approach instantiates the concept of semantic probabilistic layer (SPL) \cite{SPL}. SPL can be replaced with the predictive layer of any deep neural network. It allows to model domain knowledge as logical constraints over multiple inter-related output variables and guarantees consistency of predicted outputs. 

Given a structured input $X$, the computation of the probability of a candidate structured output $Y$ is decomposed as $p(Y|f(X)) = q_{\Theta}(Y|f(X)) \cdot c_{\mathcal{K}}(X, Y)/\mathcal{Z}(X)$, where $f(X)$ represents the feature embedding or latent representation of input $X$; $q_{\Theta}(Y|f(X))$ parameterized by $\Theta$ allows us to perform probabilistic reasoning over candidate structured outputs $Y$; 
$c_{\mathcal{K}}(X, Y)$ is a constraint reasoner to ensure that predicted structured outputs $Y$ are consistent with the domain knowledge $\mathcal{K}$ and produces evaluation $1$ (i.e., $c_{\mathcal{K}}(X, Y)$=1) only when $Y$ satisfies the declarative knowledge $\mathcal{K}$; and 
$\mathcal{Z}(x)$ is a normalization constant. 
SPL leverages the beneficial properties of probabilistic circuits \cite{PC1,PC2} for tractable learning and reasoning. It encodes 
$q_{\Theta}(Y|f(X))$ and $c_{\mathcal{K}}(X, Y)$ as circuits, and considers constrained computational graphs to achieve tractability. 
Therefore, SPL allows us to produce consistent and correct structured outputs for every inference computation task, i.e., satisfies the domain knowledge in the form of water balance equation. We refer the reader to SPL paper \cite{SPL} for additional details. 

\begin{algorithm}[t]
\caption{{Overall Training Algorithm}}
\label{alg:nsl}
\noindent\textbf{Input}: Training data of graph time-series $\mathcal{D} = \{(X_t, y_t)\}_{t=1}^{T}$; domain knowledge $\mathcal{K}$; Recurrent GNN model $f_{\theta_{fe}}$; Predictive model $F_{\theta_{p}}$; GP kernel $\kappa$ \\
\noindent\textbf{Output}: Trained feature extractor $f_{\theta_{fe}}$, predictive model $F_{\theta_{p}}$, and GP model with hyper-parameters of $\kappa$ 

\begin{algorithmic}[1] 
\STATE Train the recurrent GNN model $f_{\theta_{fe}}$ on $\mathcal{D}$ with SPL layer for domain knowledge $\mathcal{K}$ to create latent representations $z_i$=$f(X_i)$  \em{// Section 3.1}

\STATE Train the predictive regression model $F_{\theta_{p}}$ using $\calD_\tr = \{z_i, y_i\}_{i=1}^{T}$ with $T$ regression examples using the importance weighted training approach \em{// Section 3.2}
\STATE Train the GP model on $\calD_\tr = \{z_i, y_i\}_{i=1}^{T}$ to optimize the hyper-parameters of the kernel $\kappa$ \em{ // Section 3.3}
\RETURN feature extractor, predictor, and GP model
\end{algorithmic}
\end{algorithm}

\subsection{Importance Weighted Training}
\label{sec:IWtraining}

Given the learned latent representations $\{z_i=f(X_i)\}_{i=1}^T$ from the recurrent GNN model, we describe a novel importance weighted training approach to improve the generalization accuracy of the predictive model $F$. 

\vspace{0.75ex}

\noindent {\bf Notations.} Our goal is to train regression models over the training data $\calD_\tr = \{z_i, y_i\}_{i=1}^{T}$ with $T$ examples. Suppose each data sample $(z_i, y_i)$ is drawn from a target distribution $\mathcal{P}$ over the space $\mathcal{Z} \times \mathcal{Y}$ such that $z_i$ is an input from the input space $\mathcal{Z}$ and $y_i \in \mathcal{Y}$ is the corresponding ground-truth output. Let $F : \calZ \rightarrow \calY$ denote a predictive model trained on the training set $\calD_\tr$.
A loss function $\ell : \calY \times \calY \rightarrow \R$ is used to measure the accuracy of predictions made by $F$. In this paper, we consider the residual $\ell(F(z), y) = | F(z) - y |$ as loss and assume that there exists an upper bound of loss as  $L = \max_{ (z, y) \in \calZ \times \calY } \ell(F(x), y) < \infty$.
Typically, $F$ is expected to reach a small {\it true risk}, defined as $R(F) = \E_{(z,y) \sim \calP} [ \ell(F(z), y) ]$ in the sense of population.
Unfortunately, we cannot measure $R(F)$ due to the availability of only finite training samples drawn from $\calP$. 

Hence, the {\it empirical risk} $\widehat R(F) = \frac{1}{T} \sum_{i=1}^T \ell(F(z_i), y_i)$ on $\calD_\tr$ is employed to estimate it, leaving a standard estimation error bound in $O(1/\sqrt{T})$ (see Lemma \ref{lemma:generalization_bound_iw}).
The rate of convergence of generalization error bound is particularly important when the number of training samples $T$ is small. For example, to ensure the same $O(10^{-2})$ generalization error bound, $O(1/\sqrt{T})$ requires $T=O(10^4)$, while the faster rate $O(1/T)$ only needs $T=10^2$, significantly reducing the requirement of large training sets from modern ML methods.

\vspace{0.75ex}

\noindent {\bf High-level Algorithm.} Importance-weighted (IW) training is an iterative gradient-based approach. It assigns importance weight $\omega(z, y)$ to each sample $(z, y) \in \calD_{\tr}$ such that $\omega(z, y)$ is directly proportional to the loss $\ell(F(z), y)$ based on the current model $F$ and employs the weighted gradient to update the model parameters of $F$ in each iteration. 
Analogous to the empirical risk $\widehat R(F)$, we define the {\it IW empirical risk} with weights $\omega(z, y)$ for $F$ as 
\vspace{-0.06in}
\begin{align}\label{eq:empirical_IW_risk}
\widehat R_\omega (F)
=
\frac{1}{T} \sum_{i=1}^T \omega(z_i, y_i) \cdot \ell(F(z_i), y_i) .
\end{align}
$\widehat R_\omega(F)$ is general enough to include the empirical risk $R(F)$ as a special case: setting $\omega(z_i, y_i) = 1$ for all data samples reduces $\widehat R_\omega(F)$ to $\widehat R(F)$. In what follows, we show that under some configurations of importance weighting function $\omega$, $\widehat R_\omega(F)$ gives a tighter estimation of the true risk $R(F)$.  

\vspace{1.0ex}

\noindent {\bf Theoretical Analysis.} The following lemma shows the generalization error bound for IW empirical risk $R_\omega(F)$.

\begin{lemma}
\label{lemma:generalization_bound_iw}
(Error bound of IW risk, Theorem 1 in \cite{cortes2010learning})
Let $M = \sup_{z, y} \omega(z, y)$ denote the infinity norm of $\omega$ on the domain.
For given $F$ and $\delta > 0$,
with probability at least $1-\delta$, the following bound holds:
\begin{equation}\label{eq:learning_bound_IW}
R(F) - \widehat R_\omega(F)
\leq
\frac{ 2M \log(1/\delta) }{ 3T } + \sqrt{ \frac{ 2 d_2(\calP || \frac{\calP}{\omega}) \log(1/\delta) }{ T } } ,
\end{equation}
where $d_{2}(\calP \left| \right|  \mathcal{Q})$ = $\int_x \calP(x) \cdot \frac{ \calP(x) }{ \mathcal{Q}(x) } dx$ is the base-$2$ exponential for R\'enyi divergence of order $2$ between two distributions $\calP$ and $\mathcal{Q}$ and $T$ is the number of training samples.
\end{lemma}

In the above prior result, the last term on the right side ($O(\sqrt{ d_2 / T })$) typically dominates the first term ($O( M / T )$), i.e., $O( M / T ) \ll O(\sqrt{ d_2 / T })$. Consequently, the choice of $\omega$ mainly changes the quantity of R\'enyi divergence $d_2$ and thus the dominating term $O(\sqrt{ d_2 / T })$.
In what follows, we theoretically show that a simple principle to choose group-wise weights $\omega$ can easily shrink $d_2$ in the dominating term compared with the conventional empirical risk. Specifically, the grouping principle is based on the prediction loss $\ell(F(z), y)$ of $F$ overall data samples. 
Given a total number of groups $K$, denote the group $k$ by $\calD_k$ such that any data from $\calD_k$ suffers smaller loss than that from the next group $\calD_{k+1}$, i.e., $\ell(F(z), y) \leq \ell(F(z'), y')$ for all $(z, y) \in \calD_k$ and $(z', y') \in \calD_{k+1}$ with $k \in \{1, ..., K-1\}$ (see discussion in ``{\bf Practical Algorithm}'').
Accordingly, the probability of drawing data from $\calD_k$ is denoted by $\calP_k = \P\{ (z, y) \in \calD_k \}$.
Specifically, for the fixed $F$ and the same analysis framework under Lemma \ref{lemma:generalization_bound_iw}, let $\widehat B$ and $\widehat B_\omega$ be the generalization error bounds of empirical risk and IW empirical risk, respectively, i.e., 
$R(F) - \widehat R(F) \leq \widehat B$ and $R(F) - \widehat R_\omega(F) \leq \widehat B_\omega$.
Below we prove that $\widehat B_\omega < \widehat B$ under a certain group-wise setting for $\omega$. Before proceeding to the main theorem, we define group-wise weights as $\{ \omega_k \}_{k=1}^K$, where $\omega_k$ is assigned to data from group $\calD_k$ as the weight shown by $\widehat R_\omega(F)$ in (\ref{eq:empirical_IW_risk}).

\begin{theorem}
\label{theorem:improved_generalization_bound}
(IW Improves generalization bound)
Assume $\frac{ 2M \log(1/\delta) }{ 3T } \leq \sqrt{ \frac{ 2d_2(\calP || \frac{\calP}{\omega}) \log(1/\delta) }{ T } }$ in (\ref{eq:learning_bound_IW}) of Lemma \ref{lemma:generalization_bound_iw}.
Set $\omega(z, y) = \omega_k$ for $(z, y) \in \calD_k$ such that $\calP_k \cdot \omega_k = \frac{ k^a }{ K^{b+1} }$ for any $k \in \{1, ..., K\}$. Under the conditions $0 < a \leq \min\{ K-1 , b-1/\ln(K) \}$, and $K \geq 2$, the IW empirical risk $\widehat R_\omega(F)$ gives a tighter generalization bound than empirical risk $\widehat R(F)$ with $\widehat B_\omega = \sqrt{ \frac{ 2 \log(1/\delta) }{ (a+1) T } }$ and $\widehat B = \sqrt{ \frac{ 2 \log(1/\delta) }{ T } }$.
\end{theorem}

The complete proof of Theorem 1 is in the Appendix.

\vspace{0.75ex}

\noindent
{\bf Remark.}
Theorem \ref{theorem:improved_generalization_bound} shows some mild conditions to achieve a tighter generalization bound by strategically setting the group-wise weights $\omega_k$ in IW method than empirical risk, i.e., $\widehat B_\omega < \widehat B$.
Furthermore, it demonstrates how fast the hyper-parameter $a$ reduces $\widehat B_\omega$, and thus builds the connection between $a$ and the tightness of the bound.
For example, setting $a = 10^2$ makes $\widehat B_\omega$ $10$ times smaller than $\widehat B$. 

\vspace{0.75ex}

\noindent {\bf Practical Algorithm.} To devise a practical IW algorithm, we need to appropriately set the values of $K$ (no. of groups) and $\omega_k$ (weights for each of the $k$ groups). The key idea of setting the group-wise weight is to make $\omega_k \propto k / \calP_k$ when we group data into $\{\calD_k\}_{k=1}^K$ such that all data in $\calD_{k}$ suffers smaller loss than that the data in $\calD_{k+1}$.
We propose range-wise weighting, i.e., grouping data into several bins (loss intervals) according to their loss values.
For example, for $K=2$, a threshold $\tau$ partitions all data into two bins, i.e., the group suffering smaller loss $\calD_1 = \{ (z, y) : \ell(z, y) \leq \tau \}$ and the other one suffering larger loss $\calD_2 = \{ (z, y) : \ell(z, y) > \tau \}$, with $\calP_1$ and $\calP_2$ denoting the percentage of data in the two groups.
This strategy can be easily generalized to $K > 2$ by using multiple thresholds, such as uniformly grouping data into $K$ ranges with interval $\frac{L}{K}$ for maximal loss $L$:
$\calD_k = \{ (z, y): \frac{ L (k-1) }{ K } \leq \ell(F(z), y) \leq \frac{ L k }{ K } \}$ for $k \in \{1,\cdots, K\}$.
In practice, typically more data samples suffer from smaller training loss, i.e., concentrating around the true output, which implies $\calD_k$ with smaller group ID $k$ leads to larger $\calP_k$.
Hence, setting $\omega_k \propto k / \calP_k$ indicates that group $\calD_k$ with smaller group ID $k$ gets smaller weight $\omega_k$, 
In contrast, larger weights $\omega_k$ are encouraged for $\calD_k$ with larger group ID $k$. 
Training model with this IW configuration improves the generalization performance of $F$ according to Theorem \ref{theorem:improved_generalization_bound}. 

In our implementation, we employed this strategy for assigning IWs with $K$=10.

\vspace{-0.06in}
\subsection{Gaussian Process with Deep Kernels for UQ}

Gaussian processes (GPs) \cite{williams2006gaussian} are considered as gold standard for uncertainty quantification (UQ). GPs are non-parametric models and allows the practitioner to incorporate domain knowledge in the form of kernels $\kappa(x, x')$ to measure similarity between pairs of inputs $x$ and $x'$ (e.g., RBF kernel). 

GPs are Bayesian models whose posterior mean and standard deviation provide prediction and uncertainty for each input.  
A major drawback of GPs is that they have difficulty scaling to high-dimensional structured inputs (e.g., time-series of graphs) as in our problem setting. On the other hand, deep kernel learning (DKL) \cite{wilson2016deep} is a recent technique that has shown to be a promising solution towards handling this drawback. To synergistically combine the strengths of GPs and DKL, we parameterize the kernel function of the GP model with a neural network feature extractor that is passed to a canonical distance-aware kernel such as RBF kernel. Specifically, we add a GP layer on top of the latent representations learned by the recurrent GNN model. 
Given the learned latent representations $z_i=f(X_i)$ for all $i$=1 to $T$, we create a GP using $\calD_\tr = \{z_i, y_i\}_{i=1}^{T}$ with $T$ training examples. We employ the latent representations $z_i=f(X_i)$ to define a RBF kernel as shown below.
\begin{equation}
    k(z, z')=\sigma_f e^{-||z-z'||_2^2/(2\gamma^2)}.
    \label{eq:rbf}
\end{equation}

Training GPs refers to estimating the hyper-parameters of the GP kernel.
For example, the RBF kernel in equation~(\ref{eq:rbf}) has the length-scale $\gamma$ and the signal variance $\sigma_f$ hyperparameters.
To learn the hyperparameters of the kernel, we maximize the marginal likelihood of the observed data using the training data $\calD_\tr$. Another advantage of GPs for UQ is that we can update the model online based on new training examples to improve the uncertainty estimates for future predictions. 

The predictive uncertainty estimate for an input with latent representation $z$ can be computed in closed form (for the noiseless observations setting) as given below:
\begin{align}
    \sigma^2 (z) = k(z, z) - k_*^T \mathbf{K}^{-1} k_*
\end{align}
where $k_* = [k(z, z_1), k(z, z_2) \cdots k(z, z_T)]$ and $\mathbf{K}$ is the kernel matrix computed over the latent representation of training inputs. We can compute a GP driven confidence interval for input $z$ as [$\mu(z) - 2 \sigma (z)$, $\mu(z) + 2 \sigma (z)$], where $\mu(z)$ is the posterior mean prediction of the GP.

\section{Experiments and Results}

In this section, we describe our empirical results and analysis.

\subsection{Experimental Setup}

\begin{table}[H]
\begin{tabular}{c|cccccc}
\hline
\toprule
{\bf Watershed} & BR & C & CCR & F & SFC & Y\\
\hline
{\bf \#grids} & 57 & 437 & 103 & 100 & 97 & 449\\
 \hline
\toprule
\end{tabular}
\vspace{-0.125in}
\caption{\small Description of the watershed datasets. All watersheds were observed at a monthly rate from Oct. 1979 through Sep. 2014 (i.e., 420 months). Temporal features for each grid (graph node) includes incoming precipitation, max and min temperature, average temperature, wind speed, specific humidity, solar radiation, max and min relative humidity, and evapotranspiration. The environmental variables are obtained from a 4-km gridded dataset, regridded to the VIC-CropSyst model's resolution of 6-km via linear interpolation.}
\label{tab:datasets}
\end{table}

\vspace{-0.15in}
\noindent {\bf Watershed Datasets.} We collected US Geological Survey's observed monthly streamflow at the outlet of  six watersheds in the Columbia River basin of the Pacific Northwest US : Boise River (BR), Clearwater (C), Clearwater Canyon Ranger (CCR), Flathead (F), South Fork Clearwater (SFC), and Yakima (Y). Our intent was to primarily focus on "natural" watersheds with minimal human influence, and this criterion along with a constraint on the watershed size (minimum of 1800 sq. km) resulted in the first five watersheds. Additionally, we included the Yakima River basin as an example of human-influenced watershed knowing that the model performance will likely be poor. Table~\ref{tab:datasets} provides a high-level summary.  
Each watershed has different number of grids (no. of nodes in graph) and river network corresponds to the adjacency matrix of the graph. Each graph node has multiple time-varying (monthly) environmental features. 

The watersheds encompass a span of water years, commonly used in hydrology to capture the hydrological cycles, starting in October and ending in September next year. 
We employed the following training/validation/testing splits. The first 20 years of data was used for training and the next 6 years for hyper-parameter tuning. We employ the last 9 years of data for testing the generalization of various methods. 

\vspace{0.75ex}

\noindent {\bf Configuration of Predictive Algorithms.} We compare our proposed approach with several baseline methods.

\vspace{0.5ex}

\noindent {\em 1) Physics-based model (VIC-CropSyst):} We use VIC-CropSyst \cite{malek2017vic,rajagopalan2018impacts} which comprises two coupled models---the VIC model, which is a semi-distributed macroscale hydrological model \cite{liang1994simple}, and  the CropSyst model \cite{stockle2003cropsyst}. For this study, the simulations were performed with naturalized water settings---i.e., 
without extractions of water from the streams such as irrigation.

\vspace{0.5ex}

\noindent {\em 2) Recurrent Convolutional GNN (RCGNN):} We train the RCGNN model (Section 3.1) using PyTorch to optimize the Mean Squared Error (MSE) loss. It is a pure data-driven model without the knowledge of water balance equation.

\vspace{0.5ex}

\noindent {\em 3) Physics-guided RCGNN (PG-RCGNN):} This method incorporates water balance equation (WBE) into the training of RCGNN following approach by \cite{TGDS}. It employs the combined loss $\text{loss}_{mse} + \lambda_{wbe} \cdot \text{loss}_{wbe}$, where the $\text{loss}_{wbe} = max(0, \hat{y} + ET - P)$ term penalizes predictions deviating from the water balance equation.

\vspace{0.5ex}

\noindent {\em 4) RCGNN with SPL layer (RCGNN-CRL):} This approach employs constrained reasoning and learning (CRL) by adding a semantic probabilistic layer to incorporate the water balance equation in yearly aggregates. We use the Pylon software library \cite{ahmed2022pylon} for this implementation. 

\vspace{0.5ex}

\noindent {\em 5) RCGNN-CRL with Importance Weighted Training (RCGNN-CRL-IW):} This is a variant of RCGNN-CRL which uses importance weighted training described in Section 3.2. We employ $K$=10 groups and the specific weighting strategy detailed as part of the practical algorithm.

\noindent Please see the appendix section \ref{appendix:hyper_param} for hyper-parameter tuning. 

\vspace{0.75ex}

\noindent {\bf Configuration of UQ Algorithms.} We compare our proposed Gaussian process approach with a dropout baseline.

\vspace{0.5ex}

\noindent \textit{1) Gaussian process w/ deep kernel learning (GP w/ DKL).} 
We define a GP model by defining a RBF kernel over the latent representations $\{z_i$=$f(X_i)\}_{i=1}^{T}$ learned by the recurrent GNN model. We find the hyper-parameters of the RBF kernel by optimizing the negative log-likelihood of data. 

\vspace{0.5ex}

\noindent \textit{2) Dropout method}: We introduce two dropout layers after obtaining the latent representation to inject stochasticity into inference \cite{hinton2012improving}. To obtain output variance $\sigma$, we generate distinct sets of dropped neurons within each iteration, to capture the intrinsic uncertainty of deep model. The prediction interval is [$\hat{y} - 2 \sigma$, $\hat{y} + 2 \sigma $] where $\sigma$ is calculated over $30$ iterations with a dropout rate of 0.2.

\vspace{0.5ex}

\noindent {\bf Evaluation Methodology.} For comparing the predictive accuracy of different methods, we employ two different metrics: Normalized Nash–Sutcliffe Efficiency (NNSE) \cite{nossent2012application} and Mean Absolute Error (MAE). NSE is a widely recognized metric to assess the precision of hydrological models \cite{nash1970river} and is calculated as:
\begin{align}
NSE = 1 - \frac{\sum_{i=1}^n (y_i - \hat{y}_i)^2}{\sum_{i=1}^n (y_i - \bar{y})^2}
\end{align}
Here, $n$ is the number of distinct year-months during testing for the streamflow prediction, $y_i$ is the observed value at month $i$, $\hat{y}_i$ is the predicted value at month $i$, and $\bar{y}$ is the mean of the observed values. Since the theoretical lower limit of the NSE $(-\infty)$ can lead to issues in implementations, we normalize the NSE within the range of [0, 1]. The Normalized NSE is given by $NNSE$=$\frac{1}{2-NSE}$. NNSE takes into account both accuracy of predictions as well as the spread of predicted values around  ground truth. Hence, a higher NNSE value indicates better predictive performance. 
For comparing the performance of UQ, we employ two metrics: prediction interval (PI) size and marginal coverage (whether ground truth lies in the PI or not) and . There is an inherent trade-off between coverage and PI size. We can get large coverage by producing large PIs and small PIs which achieve low coverage. Ideally, we prefer to achieve some minimum target coverage (e.g., 90\%) and smaller prediction intervals. We report all results with a single training run.

\subsection{Results and Discussion}

\noindent {\bf Performance of Streamflow Predictive Models.} Table~\ref{tab:NNSE_comparison_high_flow} 
shows the accuracy results comparing different streamflow prediction models across six watersheds in the high-flow months. We make the following observations. {\bf 1)} All data-driven models perform better than physics-based model VIC-CropSyst across all watersheds. {\bf 2)} All methods that combine data and knowledge perform better than the pure data-driven model RCGNN for all cases. This demonstrates the value of domain knowledge in the form of the water balance equation  
 {\bf 3)} Importance-weighted training improves the predictive accuracy of RCGNN-CRL-IW over the standard training. 
This result demonstrates the practical value of our theoretically-justified IW training approach. {\bf 4)} Accuracy enhancement, relative to the physics-based model, varies by watershed, and time of the year (Fig ~\ref{fig:performance_high_flow}, Fig ~\ref{fig:time_series_plot} and Fig ~\ref{fig:Median_monthly_error} in Appendix).
\begin{table}[H]
\begin{tabular}{p{0.5cm}p{1.1cm}p{1.1cm}p{1.1cm}p{1.1cm}p{1.1cm}}
\hline
\textbf{} & 
{\small {\begin{tabular}[c]{@{}l@{}}VIC- \\ CropSyst\end{tabular}} }& 
{\small {RCGNN} }& 
{\small {\begin{tabular}[c]{@{}l@{}}PG- \\ RCGNN \end{tabular}}} & 
{\small {\begin{tabular}[c]{@{}l@{}}RCGNN- \\ CRL\end{tabular}}} & 
{\small \hspace{-0.1in} {\begin{tabular}[c]{@{}l@{}}RCGNN- \\ CRL-IW\end{tabular}}} \\ \hline
BR              & 0.823 & 0.915 & 0.919 & 0.93 & \textbf{0.934} \\
C               & 0.837 & 0.902 & 0.925 & 0.948 & \textbf{0.952} \\
CCR & 0.817 & 0.883 & 0.896 & 0.931 & \textbf{0.935} \\
F                 & 0.684 & 0.885 & 0.902 & 0.917 & \textbf{0.936} \\
SFC    & 0.741 & 0.834 & 0.866 & 0.91 & \textbf{0.92} \\
Y                   & 0.212  & 0.405 & 0.509 & 0.635 & \textbf{0.654} \\
\hline
\toprule
\end{tabular}
\vspace{-0.125in}
\caption{\small Comparison of predictive models across watersheds during high flow months. Values are reported in terms of the NNSE metric (higher the better).
NNSE values less than 0.5 indicate that the mean is a better predictor than the model's output.
}
\label{tab:NNSE_comparison_high_flow}
\end{table}

Table~\ref{tab:constraint_violation} illustrates the frequency and magnitude of water balance equation violations by PG-RCGNN and RCGNN-CRL-IW across various datasets during test years. Reported magnitudes for PG-RCGNN vary between datasets, reflecting the disparity in violation severity. Only years with constraint violations are included in the magnitude calculations, for a fair comparison.
Conversely, RCGNN-CRL-IW consistently maintains the water balance equation across all test years and datasets. This result demonstrates the effectiveness of RCGNN-CRL-IW in always obeying the water balance equation during inference.

\begin{table}[H]
\begin{center}
\begin{tabular}{p{1.1cm}p{0.8cm}p{0.8cm}p{0.8cm}p{0.8cm}p{0.8cm}p{0.6cm}}
\toprule
             & BR & C & CCR & F & SF & Y \\
\hline
{\small {\begin{tabular}[c]{@{}l@{}}PG-\\ RCGNN\end{tabular}} } & 
{\small {\begin{tabular}[c]{@{}l@{}}0.44\\ (51.41)\end{tabular}}} & {\small {\begin{tabular}[c]{@{}l@{}}0.44\\ (25.93)\end{tabular}}} & {\small {\begin{tabular}[c]{@{}l@{}}0.44\\ (106.1)\end{tabular}}} & {\small {\begin{tabular}[c]{@{}l@{}}0.67\\ (52.26)\end{tabular}}} & {\small {\begin{tabular}[c]{@{}l@{}}0.11\\ (8.09)\end{tabular}}} & {\small {\begin{tabular}[c]{@{}l@{}}0.0\\ (0.0)\end{tabular}}} \\
\hline
{\small {\begin{tabular}[c]{@{}l@{}}RCGNN-\\ CRL-IW\end{tabular}}} & {\small {\begin{tabular}[c]{@{}l@{}}0.0\\ (0.0)\end{tabular}}} & {\small {\begin{tabular}[c]{@{}l@{}}0.0\\ (0.0)\end{tabular} }} & {\small {\begin{tabular}[c]{@{}l@{}}0.0\\ (0.0)\end{tabular}}} &  {\small {\begin{tabular}[c]{@{}l@{}}0.0\\ (0.0)\end{tabular}}} & {\small {\begin{tabular}[c]{@{}l@{}}0.0\\ (0.0)\end{tabular}}} & {\small {\begin{tabular}[c]{@{}l@{}}0.0\\ (0.0)\end{tabular} }}\\
\hline
\end{tabular}
\vspace{-0.125in}
\caption{Comparison of fraction (magnitude) of constraint violation for PG-RCGNN and RCGNN-CRL-IW. Fraction is the proportion of test years violating the constraint, while magnitude is the average yearly constraint violation measured in millimeters.}
\label{tab:constraint_violation}
\end{center}
\end{table}

\begin{figure}[t]
    \includegraphics [width=0.5\textwidth] {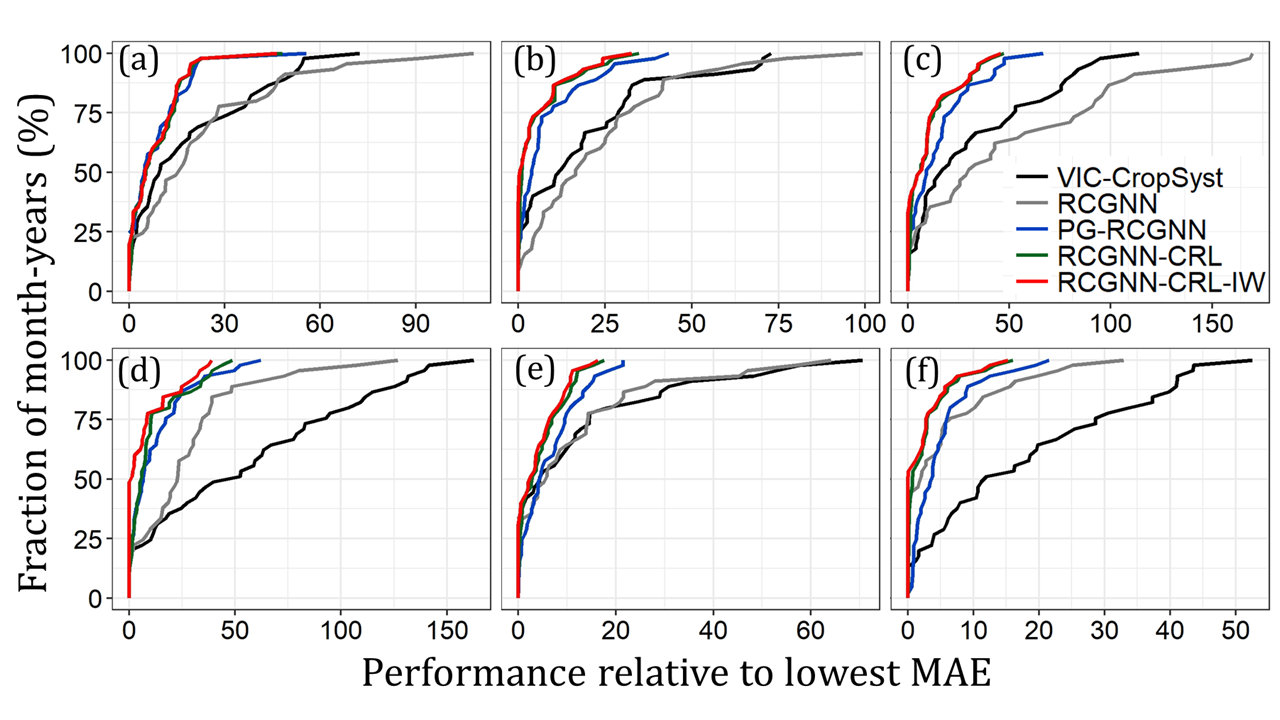}
    \caption{\small Relative performance chart comparing all models for high flow months (a) Boise, (b) Clearwater, (c) Clearwater Canyon Ranger, (d) Flathead, (e) Southfork Clearwater, and (f) Yakima. For each test month and year, each model's MAE was compared against the best performing model's MAE for that month and year (difference shown in X-axis). The fraction of test month-years for which that performance is achieved is shown along the Y-axis. Closer a curve is aligned with the Y-axis for larger fraction, the better.}\label{fig:performance_high_flow}
\end{figure}

Fig~\ref{fig:performance_high_flow} shows the relative performance of each model compared to other models. RCGNN-CRL-IW (red line) yields the best performance most consistently for the high flow months. Even, when it is not the best model, its degradation in MAE relative to the best model is low (i.e., the red line sticks closer to the Y-axis. On the other hand, the VIC-CropSyst and RCGNN models are best performing in very few cases, and when they are not the best models, the degradation in performance is relatively high (the lines are further away from the Y-axis). This is primarily due to the failure to capture the timing and magnitude of peak  streamflow (Fig~\ref{fig:time_series_plot} ). 

\vspace{0.5ex}

\noindent {\bf Performance of UQ Methods.} Table~\ref{tab:UQ_comparison} shows the UQ performance of GP w/ deep kernel learning (DKL) and dropout. The prediction intervals are in streamflow units. We make the following observations. {\bf 1)} Dropout produces very small prediction intervals but the coverage is low (less than 80 percent for all but one watershed) which is not desirable. {\bf 2)} GP w/ DKL achieves significantly higher coverage (more than 91 percent for all watersheds) at the expense of a slight increase in the prediction interval size over the dropout method.  
\begin{table}[t]
\begin{center}
\begin{tabular}{l|cc|cc}
\multirow{2}{*}{} & \multicolumn{2}{c}{Dropout} & \multicolumn{2}{c}{GP w/ DKL} \\
                  & coverage     & PI width     & Coverage   & PI width  \\
\hline
BR & 74.07 & 4.77 & 91.66 & 5.99 \\
C & 76.85 & 4.7 & 93.52 & 6.3 \\
CCR & 79.62 & 5.47 & 92.59 & 6.27 \\
F & 75.92 & 4.92 & 94.44 & 6.28 \\
SFC & 78.7 & 4.59 & 95.73 & 6.08 \\
Y & 81.48 & 4.37 & 94.44 & 5.84 \\
\end{tabular}
\vspace{-0.125in}
\caption{Comparison of marginal coverage and prediction interval
size for dropout and GP w/ DKL approaches.}
\label{tab:UQ_comparison}
\end{center}
\end{table}

\vspace{0.5ex}

\noindent {\bf Modeling streamflow in human-influenced watersheds.} 
The VIC-CropSyst model simulates ``natural'' flows without human interventions such as reservoir operations or irrigation. Hence, its performance tends to be poor in human-influenced watersheds such as the Yakima River basin, where much river water is diverted for agriculture. The RCGNN-CRL-IW model shows better performance in such contexts, capturing some human influence nuances. Nonetheless, it exhibits relatively lower NNSE values, highlighting the challenge of modeling streamflow in human-influenced watersheds. 
This underscores the necessity to explicitly incorporate elements of human influence into the learning approach.

\vspace{-1.0ex}

\section{Roadmap to Deployment}
The intended application base for our streamflow prediction methodology are federal and state/regional agencies that manage reservoirs or engage in other forms of water resources management and planning. However, to enable such partnerships in the near term, several improvements and extensions have been planned. First, the problem formulation needs to be extended to a forecast setting where expectations of meteorological and other input variables inform the streamflow predictions. Datasets for expectations of weather are available in the same format as our current inputs and can be directly integrated.  Our current prediction accuracies provide an upper bound which will diminish when errors associated with expectations of input variables are added in. To this end, the provision of uncertainty bounds supported by our framework will be crucial for the end-user. User interfaces that display the uncertainty metrics in a way that is most useful for decision-making are needed for deployment. We have been exploring the current decision-making interfaces used by the agency to seamlessly integrate. Finally, our stakeholder agency identified  explainability of the model as a key feature of interest in transitioning to deployment and we can adapt existing work on explainable AI for this purpose.

\vspace{-1.0ex}

\section{Summary and Future Work}

This paper studied a constrained reasoning and learning approach for streamflow prediction by combining domain knowledge with deep temporal models, and Gaussian process with deep kernel learning approach for uncertainty quantification. Our experimental results demonstrated the effectiveness of our overall methodology on diverse real-world watershed datasets and in guaranteeing that constraints are satisfied for every test-time prediction. One unexplored but important challenge is to model and reason about the human influence on water availability to further improve accuracy. 

\newpage

\noindent {\bf Acknowledgements.} 
This research was supported in part by United States Department of Agriculture (USDA) NIFA award No.
2021-67021-35344 (AgAID AI Institute), and by the USDA NIFA Award \#1016467 under the Water for Agriculture program. 

\bibliographystyle{named}
\bibliography{ijcai24}

\newpage

\appendix
\onecolumn
\section{Appendix: Recurrent Convolutional Graph Neural Network}

\begin{figure}[H]
   \centering
   \includegraphics[width=0.4\textwidth]{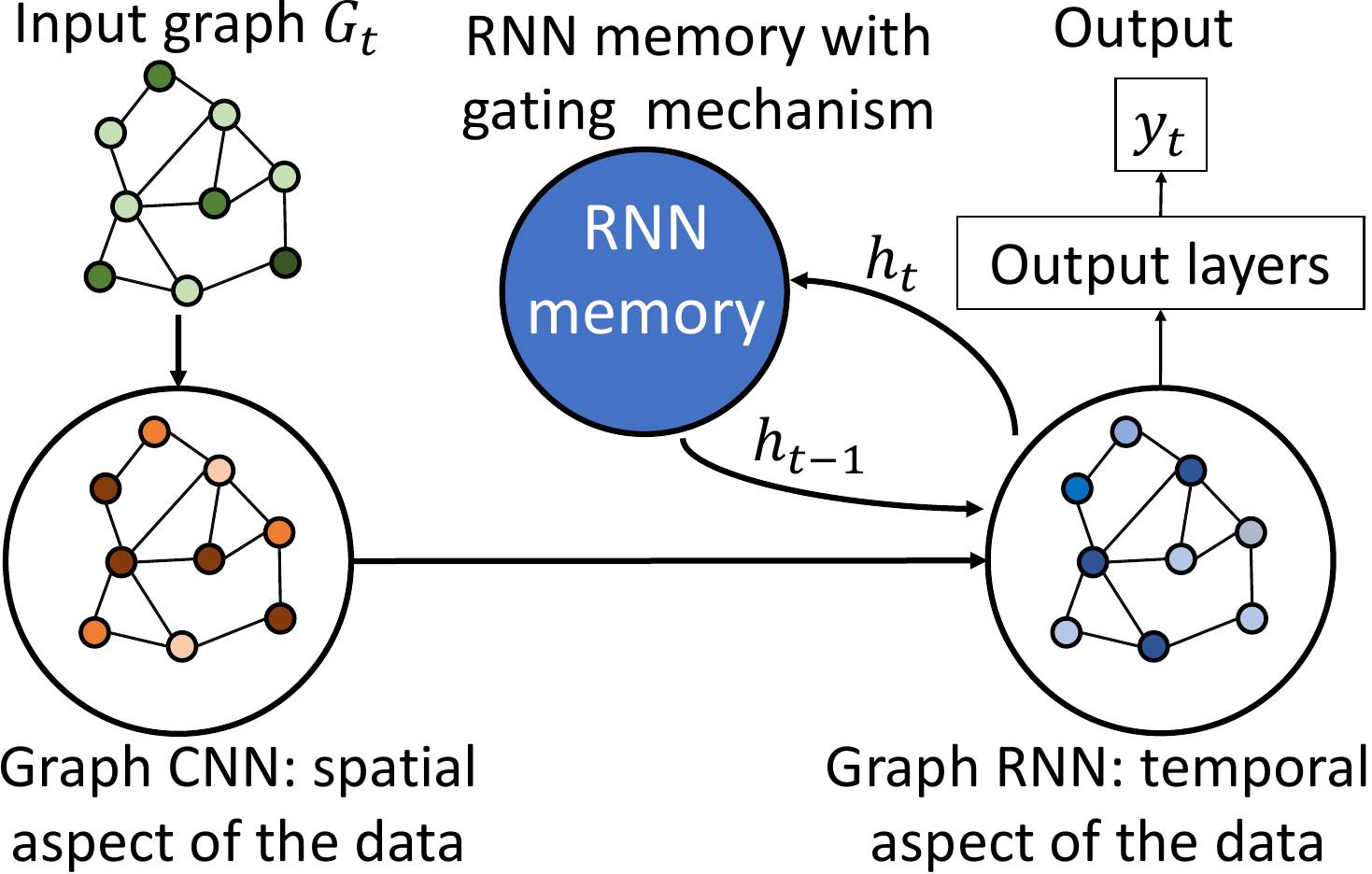}
 \vspace{-0.125in}
   \caption{Conceptual illustration of RCGNN architecture.}
   \label{fig:RCGNN}
 \end{figure}
 \vspace{-0.1in}
Figure~\ref{fig:RCGNN} depicts the RCGNN model, an architecture that captures insights from both spatial and temporal dimensions.

\section{Appendix: Proof for theorem 1}

\section*{ Proof for Theorem \ref{theorem:improved_generalization_bound} }

\begin{theorem}
\label{theorem:appendix:improved_generalization_bound}
(Theorem \ref{theorem:improved_generalization_bound} restated: IW Improves generalization bound)
Assume $\frac{ 2M \log(1/\delta) }{ 3T } \leq \sqrt{ \frac{ 2d_2(\calP || \frac{\calP}{\omega}) \log(1/\delta) }{ T } }$ in (\ref{eq:learning_bound_IW}) of Lemma \ref{lemma:generalization_bound_iw}.
Set $\omega(z, y) = \omega_k$ for $(z, y) \in \calD_k$ such that $\calP_k \cdot \omega_k = \frac{ k^a }{ K^{b+1} }$ for any $k \in \{1, ..., K\}$. Under the conditions $0 < a \leq \min\{ K-1 , b-1/\ln(K) \}$, and $K \geq 2$, the IW empirical risk $\widehat R_\omega(F)$ gives a tighter generalization bound than empirical risk $\widehat R(F)$ with $\widehat B_\omega = \sqrt{ \frac{ 2 \log(1/\delta) }{ (a+1) T } }$ and $\widehat B = \sqrt{ \frac{ 2 \log(1/\delta) }{ T } }$.
\end{theorem}

\begin{proof}
(of Theorem \ref{theorem:improved_generalization_bound})
Recall the base-$2$ exponential of R\'enyi divergence of order $2$ between $\calP$ and $\frac{\calP}{\omega}$:
\begin{align*}
d_2( \calP || \frac{\calP}{\omega} )
=
\int_{(z,y)} \calP(z, y) \cdot \frac{ \calP(z, y) }{ \frac{\calP(z, y)}{\omega(z, y)} } d(z, y)
=
\int_{(z, y)} \calP(z, y) \cdot \omega(z, y) d(z, y) .
\end{align*}
It it worth noting a special case of $\omega(z, y) = 1$ reduces IW empirical risk $R_\omega(F)$ back to empirical risk $R(F)$:
\begin{align*}
d_2( \calP || \frac{\calP}{\omega} )
=
d_2( \calP || \frac{\calP}{1} )
=
\int_{(z, y)} \calP(z, y) \cdot 1 d(z, y) 
= 1 .
\end{align*}

Then, to prove the main result regarding generalization bounds, under the assumption that $\frac{ 2M \log(1/\delta) }{ 3T } \leq \sqrt{ \frac{ 2d_2(\calP || \frac{\calP}{\omega}) \log(1/\delta) }{ T } }$ in the same analysis result (\ref{eq:learning_bound_IW}) of Lemma \ref{lemma:generalization_bound_iw}, we have the following inequalities:
\begin{align}\label{eq:gen_error_empirical_risk_vs_IW}
R(F) - \widehat R(F)
\leq &
2 \sqrt{ \frac{ 2 d_2(\calP || \calP ) \log(1/\delta) }{ T } }
= 
2 \sqrt{ \frac{ 2 \log(1/\delta) }{ T } }
=
\widehat B ,
\nonumber\\
R(F) - \widehat R_\omega(F)
\leq &
2 \sqrt{ \frac{ 2 d_2(\calP || \frac{\calP}{\omega} ) \log(1/\delta) }{ T } }
= 
\widehat B_\omega .
\end{align}

To show that IW empirical risk $\widehat R_\omega(F)$ gives a tighter generalization bound than empirical risk, it suffixes to prove that $d_2(\calP || \frac{\calP}{\omega}) \leq d_2(\calP || \calP) = 1$.
Now we divide $d_2(\calP || \frac{\calP}{\omega})$ that involves all possible data into $K$ groups as follows
\begin{align*}
d_2( \calP || \frac{\calP}{\omega} )
= &
\int_{(z,y)} \calP(z,y) \cdot \Bigg( \sum_{k=1}^K \I[ (z,y) \in \calD_k ] \Bigg) \cdot \omega(z,y) d(z,y) 
\\
= &
\sum_{k=1}^K \int_{(z,y)} \calP(z,y) \cdot \I[ (z,y) \in \calD_k ] \cdot \omega(z,y) d(z,y)
\\
\stackrel{ (a) }{ = } &
\sum_{k=1}^K \int_{(z,y)} \calP(z,y) \cdot \I[ (z,y) \in \calD_k ] \cdot \omega_k d(z,y)
= 
\sum_{k=1}^K \calP_k \cdot \omega_k ,
\end{align*}
where the above equality $(a)$ is due to the setting of group-wise important weights $\omega(z, y) = \omega_k$ for $(z, y) \in \calD_k$.
Then under the setting of group-wise weights $\omega_k$ such that $\calP_k \cdot \omega_k = \frac{ k^a }{ K^b } \cdot \frac{ 1 }{ K }$ with $a > 0$ a for all $k \in [K]$, we can further upper bound $d_2$ as follows:
\begin{align}\label{eq:ub_d_2_IW}
d_2(\calP || \frac{\calP}{\omega})
= &
\sum_{k=1}^K \calP_k \cdot \omega_k
=
\sum_{k=1}^K \frac{ k^a }{ K^b } \cdot \frac{ 1 }{ K }
\leq 
\frac{ \int_{k=1}^{K+1} k^a dk }{ K^{b+1} }
\leq 
\frac{ (K+1)^{(a+1)} }{ (a+1) K^{b+1} }
= 
\frac{ 1 }{a+1} \cdot \bigg( \frac{ K + 1 }{ K } \bigg)^{a+1} \cdot K^{a-b}
\nonumber\\
= &
\frac{1}{a+1} \cdot \bigg( 1 + \frac{1}{K} \bigg)^{a+1} \cdot K^{a-b}
\leq 
\frac{1}{a+1} \cdot \exp\bigg(\frac{a+1}{K}\bigg) \cdot K^{a-b}
\leq 
\frac{1}{a+1} \cdot \exp(1) \cdot K^{a-b}
\leq 
\frac{1}{a+1} 
,
\end{align}
where the first inequality is due to the left rule of Riemann sum for the monotonically increasing function $k^a$ with $a > 0$, 
the third inequality is due to $1+x \leq \exp(x)$,
the fourth inequality is due to the condition $a \leq K-1$,
and the last inequality is due to the condition $a \leq b - 1 / \ln(K)$.

As can be seen, in (\ref{eq:gen_error_empirical_risk_vs_IW}), $\widehat B_\omega < \widehat B$ holds:
\begin{align*}
\widehat B_\omega
=
2 \sqrt{ \frac{ 2 d_2(\calP || \frac{\calP}{\omega}) \log(1/\delta) }{ T } }
\leq 
2 \sqrt{ \frac{ 2 \log(1/\delta) }{ (a+1) T } }
<
2 \sqrt{ \frac{ 2 \log(1/\delta) }{ T } }
=
\widehat B ,
\end{align*}
where the inequality is due to (\ref{eq:ub_d_2_IW}) and the condition $0 < a$.
This immediately shows that IW empirical risk $\widehat R_\omega(F)$ gives a tighter generalization bound than empirical risk $\widehat R(F)$.
\end{proof}

\section{Appendix: Additional Experimental Results}

\begin{figure}[h!]
\centering
    \includegraphics[width=1\textwidth] {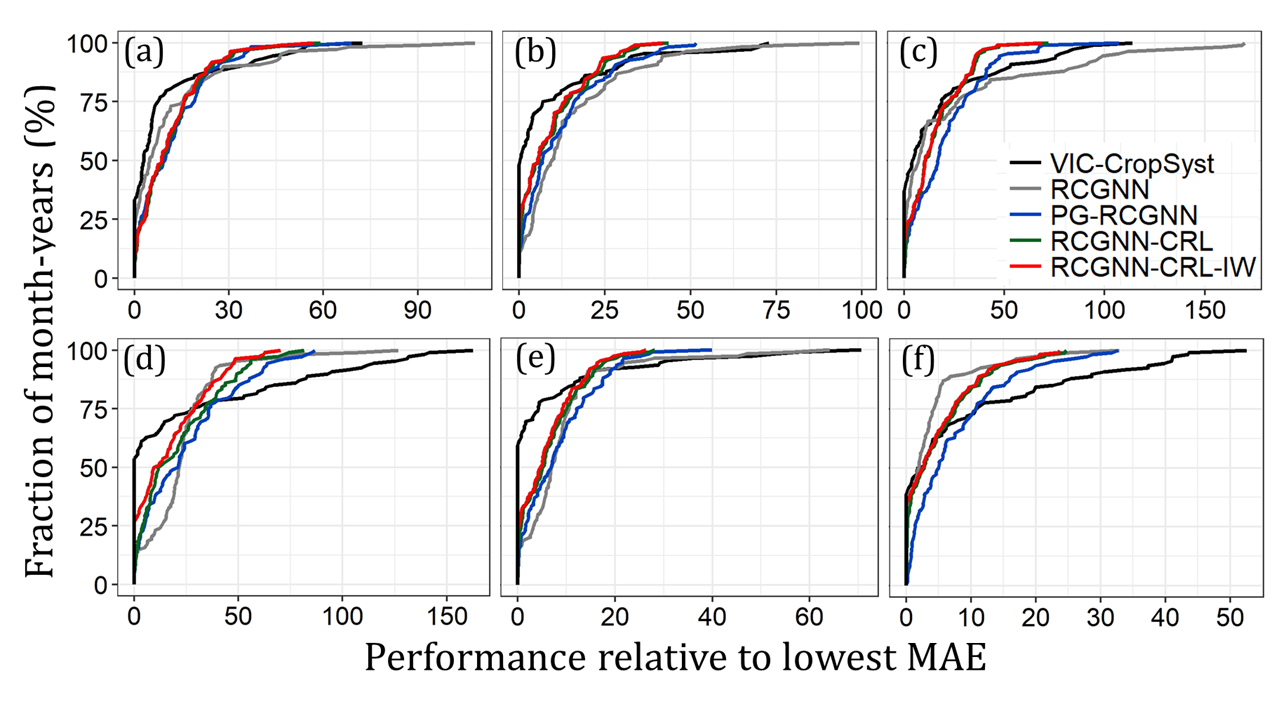}
    \caption{{\bf Additional analysis for both low- and high-flow months:} Relative performance chart comparing all models for  (a) Boise, (b) Clearwater, (c) Clearwater Canyon Ranger, (d) Flathead, (e) Southfork Clearwater, and (f) Yakima. For each test month and year, each model's MAE was compared against the best performing model's MAE for that month and year (difference shown in X-axis). The fraction of test month-years for which that performance is achieved is shown along the Y-axis. Closer a curve is aligned with the Y-axis for larger fraction, the better.}\label{fig:Rel_perf_overall}
\end{figure}

\begin{figure}[h!]
\centering
    \includegraphics[width=1\textwidth] {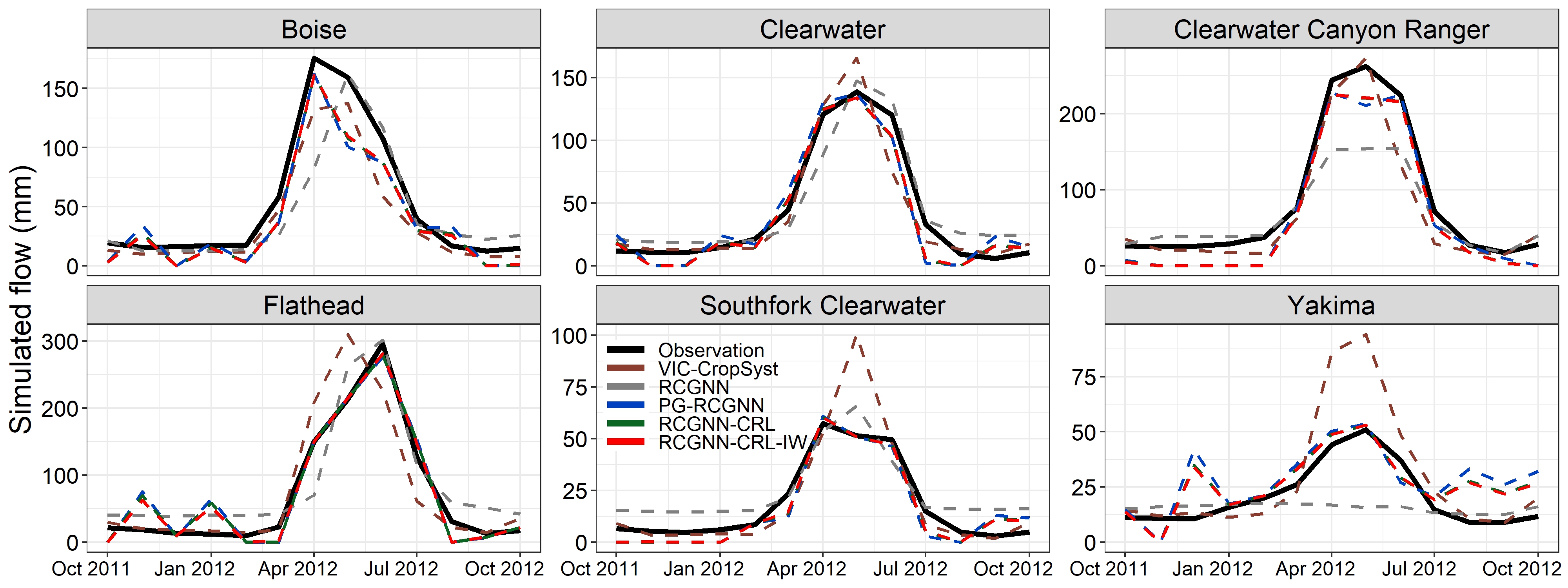}
    \caption{Time-series plot of simulated and observed streamflow for the water year 2011. }\label{fig:time_series_plot}
\end{figure}

\begin{figure}[h!]
\centering
    \includegraphics[width=1\textwidth] {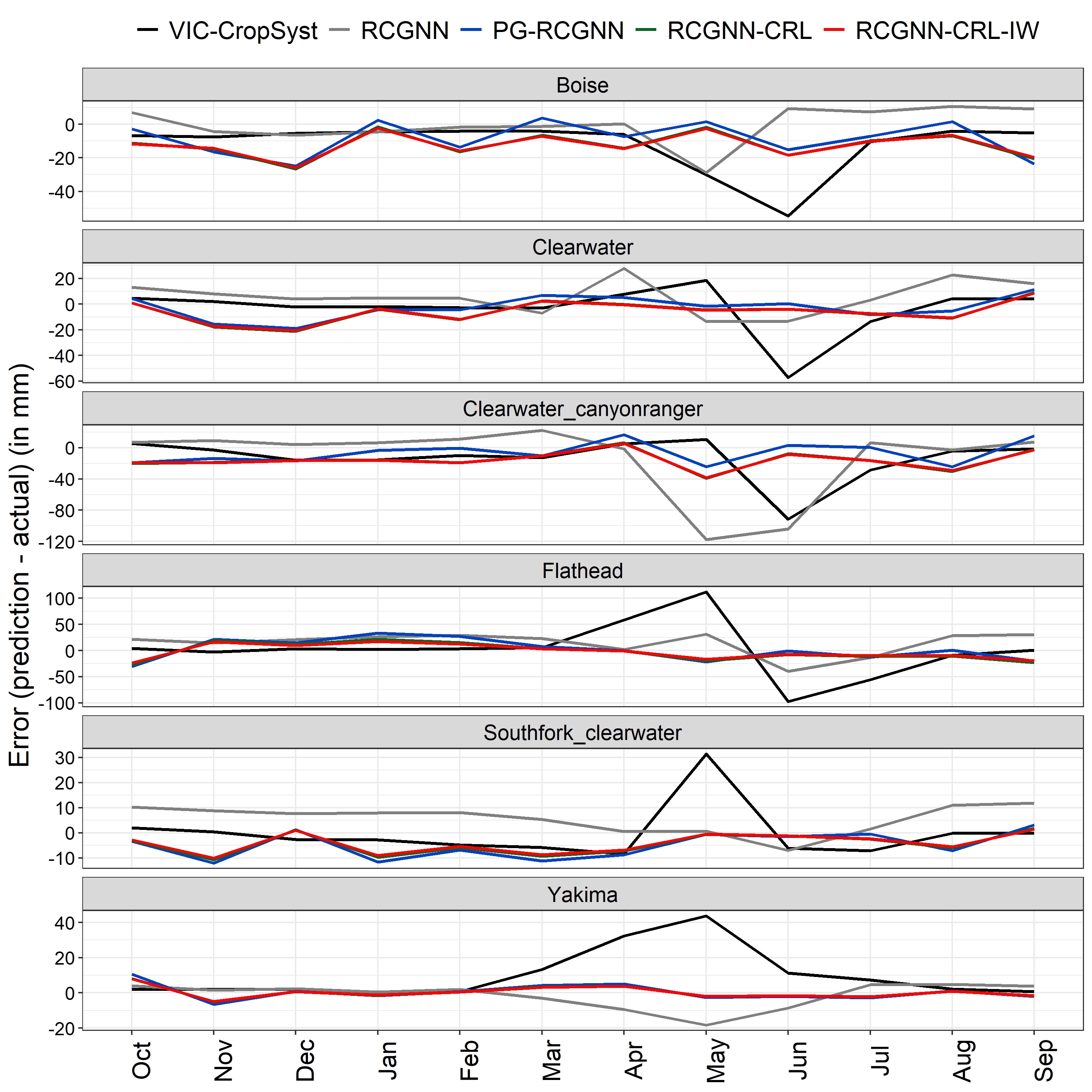}
    \caption{Median of monthly error (prediction - actual) for different watersheds. A positive value of error indicates the overestimation of simulated flow by model and a negative value indicates under-estimation}\label{fig:Median_monthly_error}
\end{figure}

 Figure~\ref{fig:Rel_perf_overall} shows the relative performance of the different models based on their performance over all the test months and years. From an overall perspective, the VIC-CropSyst model yields the best performance in the majority of the month-year cases, but as highlighted in the main paper Fig 2., it fails to perform well in high-flow months. This is also  demonstrated in Figure~\ref{fig:time_series_plot} and Figure~\ref{fig:Median_monthly_error}. Accurately capturing flows during these high-flow  months is crucial for our application context because the estimated inflows directly affect reservoir operations and are the primary focus of this paper. Figure~\ref{fig:time_series_plot} shows that the VIC-CropSyst and RCGNN model fails to capture the peak of the streamflow that leads to higher error in the flow during these months (Figure~\ref{fig:Median_monthly_error}). In addition, in some watersheds, we observe the shift in the timing of the peak streamflow (e.g., Flathead) that degrades the performance of the VIC-CropSyst and RCGNN models. While other data-driven models with included knowledge about the constraints perform better.

\begin{table}[H]
\begin{center}
\begin{tabular}{p{0.5cm}p{1.1cm}p{1.1cm}p{1.1cm}p{1.1cm}p{1.1cm}}
\hline
\textbf{} & 
{\small {\begin{tabular}[c]{@{}l@{}}VIC- \\ CropSyst\end{tabular}} }& 
{\small {RCGNN} }& 
{\small {\begin{tabular}[c]{@{}l@{}}PG- \\ RCGNN \end{tabular}}} & 
{\small {\begin{tabular}[c]{@{}l@{}}RCGNN- \\ CRL\end{tabular}}} & 
{\small \hspace{-0.1in} {\begin{tabular}[c]{@{}l@{}}RCGNN- \\ CRL-IW\end{tabular}}} \\ \hline
BR & 0.839 & 0.842 & 0.8443 & 0.8578 & \textbf{0.8651} \\
C &0.8551 &0.8536 &0.8676 &0.9018 & \textbf{0.9101} \\
CCR &0.8324 &0.8412 &0.851 &0.9058 & \textbf{0.9107} \\
F  &0.7133 &0.7937 &0.8218 &0.855 &\textbf{0.8867} \\
SFC &0.7692 &0.7659 &0.81 &0.8578 &\textbf{0.8723} \\ 
Y  &0.2315 &0.2873 &0.3798 &0.5073 &\textbf{0.5277} \\ \hline
\toprule
\end{tabular}
\vspace{-0.01in}
\caption{\small Comparison of predictive models across watersheds during all months. Values are reported in terms of the NNSE metric (higher the better).
NNSE values less than 0.5 indicate that the mean is a better predictor than the model's output.
}
\label{tab:NNSE_comparison_overall}
\end{center}
\end{table}

Table~\ref{tab:NNSE_comparison_overall} underscores the superior performance of RCGNN-CRL-IW over all other baselines, even when evaluating across all months. These results align with the findings drawn from comparing model performance specifically during high flow months, as illustrated in Table~\ref{tab:NNSE_comparison_high_flow}. This table provides supplementary details regarding the evaluation of model performance discussed in the Table~\ref{tab:NNSE_comparison_high_flow}. While the primary focus of the paper is on high flow months, overall results are also included here for comprehensive understanding. It is noteworthy that the overall performance, as presented in this appendix, exhibits slight variations as we go from high flow months to all months.

\begin{table}[H]
\begin{center}
\begin{tabular}{llll}
\toprule
 & PG-RCGNN & RCGNN-CRL & RCGNN-CRL-IW \\
 \hline
BR & 0.8192 & 0.8242 & 0.8432\\
C & 0.8471 & 0.8588 & 0.8713\\
CCR & 0.8288 & 0.8461 & 0.8571\\
F & 0.7563 & 0.7703 & 0.7835\\
SF & 0.7622 & 0.7811 & 0.7986\\
Y & 0.3675 & 0.3851 & 0.4368\\
 \hline 
\end{tabular}
\caption{\small Comparison of predictive models across watersheds using bi-weekly data. Values are reported in terms of the NNSE metric (higher the better).
NNSE values less than 0.5 indicate that the mean is a better predictor than the model's output.}
\label{tab:NNSE_biweekly}
\end{center}
\end{table}

Table~\ref{tab:NNSE_biweekly} presents the NNSE comparison for the models utilizing bi-weekly data, highlighting the effectiveness of RCGNN-CRL-IW even within a lower granularity framework. Additionally, the table reveals a marginal decrease in overall performance across all models as we go from monthly to bi-weekly datasets.

\begin{table}[H]
\begin{center}
\begin{tabular}{llll}
\toprule
 & PG-RCGNN & RCGNN-CRL & RCGNN-CRL-IW \\
 \hline
BR & 0.8136 &0.8204 &0.8317\\ 
C &0.8337 &0.8396 &0.8526\\
CCR &0.8053 &0.8178 &0.8382\\
F &0.7349 &0.7461 &0.7512\\
SF &0.7472 &0.7522 &0.7639\\
Y &0.3418 &0.3698 &0.3943\\
 \hline 

\end{tabular}
\caption{\small Comparison of predictive models across watersheds using weekly data. Values are reported in terms of the NNSE metric (higher the better).
NNSE values less than 0.5 indicate that the mean is a better predictor than the model's output.
}
\label{tab:NNSE_weekly}
\end{center}
\end{table}

Table~\ref{tab:NNSE_weekly} reaffirms the dominance of RCGNN-CRL-IW when utilizing weekly observations, outperforming other models. Moreover, we observe a consistent trend of decreasing performance as granularity decreases.

\begin{table}[H]
\begin{center}
\begin{tabular}{l|cc|cc}
\multirow{2}{*}{} & \multicolumn{2}{c}{high flow months} & \multicolumn{2}{c}{all months} \\
                  & {\small {\begin{tabular}[c]{@{}c@{}}PG- \\ RCGNN\end{tabular}} }     & {\small {\begin{tabular}[c]{@{}c@{}}PG- \\ RCGNN-IW\end{tabular}} }      & {\small {\begin{tabular}[c]{@{}c@{}}PG- \\ RCGNN\end{tabular}} }  & {\small {\begin{tabular}[c]{@{}c@{}}PG- \\ RCGNN-IW\end{tabular}} }   \\
\hline
BR & 0.919 & 0.924 & 0.844 & 0.853 \\
C & 0.925 & 0.931 & 0.868 &  0.879\\
CCR & 0.896 & 0.905 & 0.851 &  0.875\\
F & 0.902 & 0.911 & 0.822 &  0.846\\
SFC & 0.866 & 0.883 & 0.81 &  0.832\\
Y & 0.509 & 0.542 & 0.38 &  0.407\\
\end{tabular}
\caption{\small Comparison of PG-RCGNN and PG-RCGNN-IW (PG-RCGNN with importance weighted training) across watersheds during high flow and overall months. Values are reported in terms of the NNSE metric (higher the better).
NNSE values less than 0.5 indicate that the mean is a better predictor than the model's output.}
\label{tab:NNSE_PG_IW}
\end{center}
\end{table}

Table~\ref{tab:NNSE_PG_IW} shows that the importance weighted training contributes to the performance of PG-RCGNN in both high flow months and all months. This further showcases the generalizability and effectiveness of IW training in improving performance.

\section{Hyper-parameter Tuning} \label{appendix:hyper_param}
We employed the validation data to select the 
hyper-parameters for all deep neural network approaches (RCGNN, PG-RCGNN, RCGNN-CRL, RCGNN-CRL-IW). 
We selected a learning rate of 0.001, a weight decay of 0.0005, a hidden dimension of 256 for the predictive layers, and a dropout rate of 0.2.

\end{document}